\pdfoutput=1

\documentclass[11pt]{article}

\usepackage[final]{acl}

\usepackage{times}
\usepackage{latexsym}

\usepackage[T1]{fontenc}

\usepackage[utf8]{inputenc}

\usepackage{microtype}

\usepackage{inconsolata}

\usepackage{graphicx}
\usepackage{multirow}
\usepackage{booktabs}
\usepackage{cleveref}
\usepackage{algpseudocode}
\usepackage{svg}
\usepackage{tabularx}

\definecolor{CBBlue}{HTML}{56B4E9}     
\definecolor{CBOrange}{HTML}{E69F00}   

\title{Fine-Tuning on Diverse Reasoning Chains Drives \\Within-Inference CoT Refinement in LLMs}

\author{Haritz Puerto$^{1}$, Tilek Chubakov$^{1}$\\ \textbf{Xiaodan} \textbf{Zhu}$^{2}$, \textbf{Harish} \textbf{Tayyar} \textbf{Madabushi}$^{3}$, \textbf{Iryna} \textbf{Gurevych}$^{1}$\\
$^{1}$Ubiquitous Knowledge Processing Lab (UKP Lab), TU Darmstadt and\\
 ATHENE National Research Center for Applied Cybersecurity, Germany \\
$^{2}$Dept. of ECE \& Ingenuity Labs Research Institute, Queen's University, Canada\\$^{3}$University of Bath, UK\\ \href{https://www.ukp.tu-darmstadt.de}{https://www.ukp.tu-darmstadt.de}
}

\begin{document}
\maketitle
\begin{abstract}
Requiring a large language model (LLM) to generate intermediary reasoning steps, known as Chain of Thought (CoT), has been shown to be an effective way of boosting performance. Previous approaches have focused on generating multiple independent CoTs, combining them through ensembling or other post-hoc strategies to enhance reasoning. In this work, we introduce a novel approach where LLMs are fine-tuned to generate a sequence of Diverse Chains of Thought (DCoT) within a \emph{single inference step}, which is fundamentally different from prior work that primarily operate on parallel CoT generations. DCoT allows LLMs to gain the ability to perform within-inference refinement of reasoning chains without requiring external feedback. Through a rigorous set of experiments spanning a wide range of tasks that require various reasoning types, we show that fine-tuning on DCoT improves performance over the CoT baseline across model families and scales (1.3B to 70B). These improvements are particularly impactful for tasks with a large result state space, such as those involving numeric answers. Our work is also significant because both quantitative analyses and manual evaluations reveal the observed gains stem from the models' ability to refine an initial reasoning chain by generating a second, improved chain within the same inference step, demonstrating previously elusive self-improvement.
Our code and data are publicly available.\footnote{\href{https://github.com/UKPLab/acl2025-diverse-cot}{https://github.com/UKPLab/acl2025-diverse-cot}}

\end{abstract}

\section{Introduction}
Chain of Thought (CoT; ~\citealt{NEURIPS2022_9d560961}), the prompting method to generate intermediate reasoning steps to answer a question, is recognized as a simple yet effective mechanism for improving the performance of large language models (LLMs).
Given that requiring models to generate intermediary steps improves performance, it stands to reason that requiring models to generate multiple chains could further improve performance. Prior work exploring this idea includes that by \citet{wang2023selfconsistency}, wherein they generate multiple CoTs and ensemble them with a voting mechanism. 
However, these approaches, and others like them (see Section \ref{sec:related-works}), rely on independently generated CoTs, which prevents the model from accessing previously generated chains during inference. This independence limits the potential for within-inference refinement and the ability to build upon earlier reasoning paths.

\begin{figure}[t]
\centering
\includegraphics[width=0.47\textwidth]{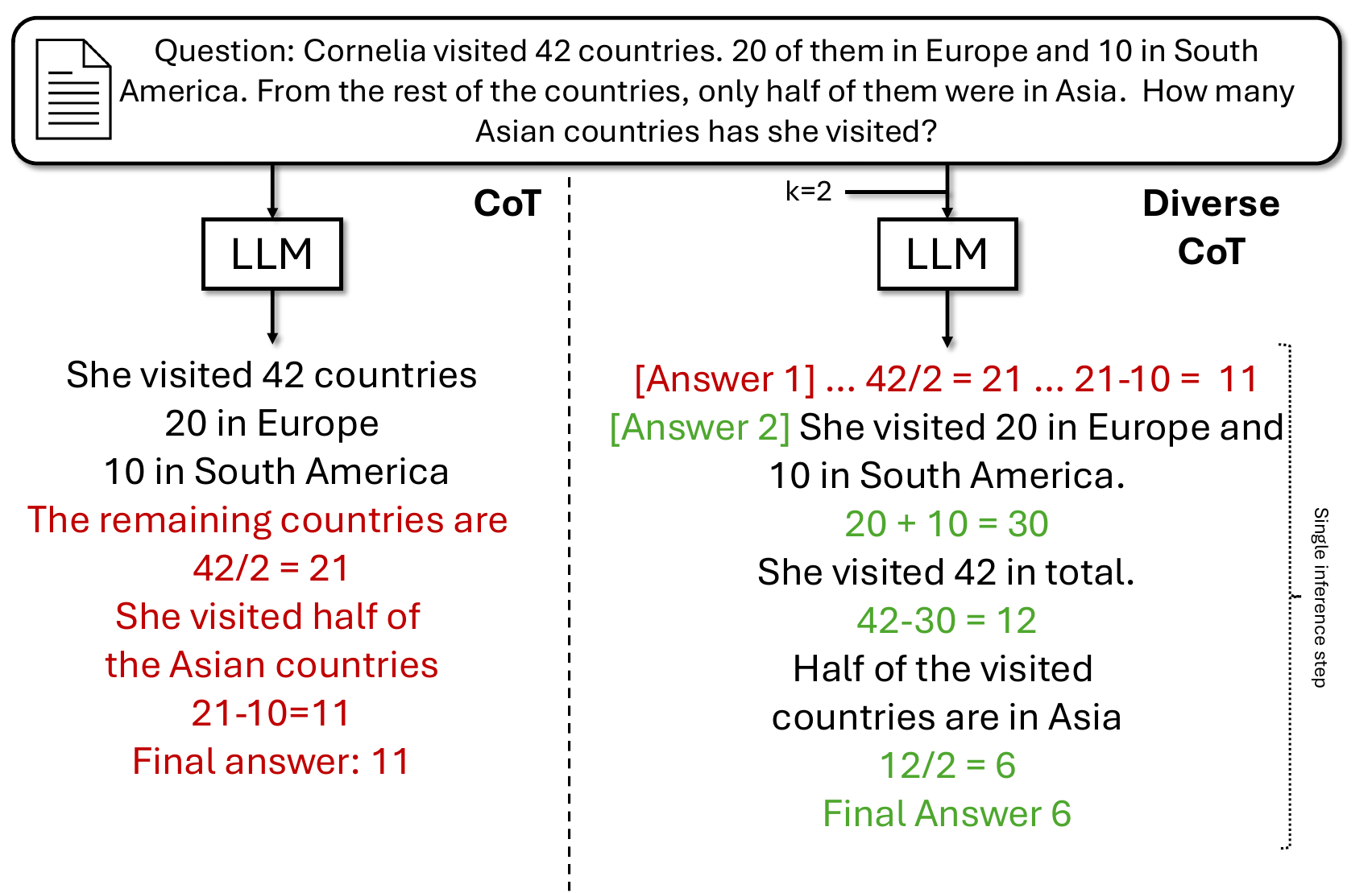}
\caption{Diverse CoT ($k=2$) generates $k$ CoTs in a single inference step and selects the correct answer.}
\label{fig:intro}
\end{figure}

Therefore, we present a training method that enables LLMs to generate multiple diverse reasoning chains sequentially in \textit{a single inference step}. 
Through our experiments, we find that this method drives performance gains by allowing the model to refine subsequent CoTs based on earlier ones.
To achieve this, we construct a training dataset of Diverse CoTs (DCoT), where a single question is associated with multiple valid CoTs. While previous studies treated each (question, CoT) pair as a standalone data point \citep{ho-etal-2023-large, huang-etal-2023-large}, we propose to concatenate CoTs into a single sequence, forming training pairs of the format (question, [CoTs]). In this way, models learn to generate multiple CoTs in a single inference step. While all our training CoTs represent correct reasoning chains, we hypothesize that this training regime will enable LLMs to generate sequentially better CoTs---up to a certain number of iterations---as they have access to previous reasoning chains, leading to improved performance. 

We demonstrate that fine-tuning using DCoTs improves LLM performance over the CoT baseline by rigorously testing on a range of tasks requiring different types of reasoning across model families and scales (1.3B to 70B). We further identify the subset of tasks, namely those with a large result state space, such as those involving numeric answers, to particularly benefit from our method.
In addition, we show that generating a single CoT on the DCoT fine-tuned model yields comparable results to the CoT baseline, while generating two or more CoTs yields clear gains on average across all tasks for all models. This, along with the manual evaluation that we present below, demonstrates that the gains provided by DCoT arise not from random perturbations but from iterative refinement within a single inference step.
Additionally, we show that DCoT fine-tuned models can be further augmented by the same methods that boost CoT, such as the self-ensemble of CoTs \citep{NEURIPS2022_9d560961}. These results suggest that instruction-tuning datasets can be easily augmented with DCoT data, given that many existing datasets already include CoT examples, often with multiple CoTs per question \citep{ho-etal-2023-large, huang-etal-2023-large}. This makes the creation of DCoT training datasets both practical and efficient.
Specifically, the contributions of this work are as follows: 
\begin{itemize}
    \itemsep0em 
    \item We introduce a novel method that fine-tunes LLMs to generate multiple reasoning chains within a single inference step, enhancing subsequent chains and boosting performance.
    \item We rigorously demonstrate the effectiveness of our method on a range of LLM families and sizes across multiple reasoning tasks, identifying task types where it performs best—those with large result state spaces.
    \item Through a combination of empirical and manual analysis, we show that DCoT achieves gains through within-inference revision of its first CoT without external feedback or prompt optimization, which, to the best of our knowledge, is the first work to do so.
\end{itemize}

\begin{figure*}[t]
\centering
\includegraphics[width=0.99\textwidth]{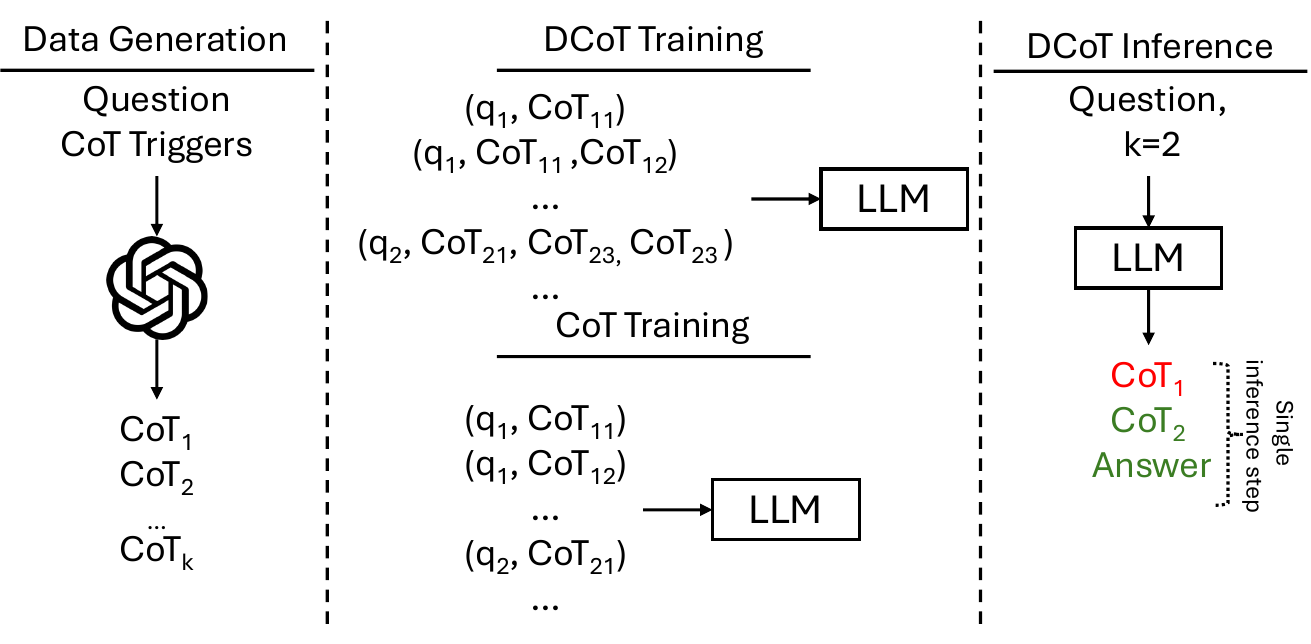}
\caption{We train on a series of CoTs to make the model learn how to generate multiple CoTs in one inference step. \texttt{DCoT} and \texttt{CoT} have the same amount of CoTs. However, \texttt{DCoT} is trained with different amounts of $k$ CoTs for a given query. At inference time, users can pick any $k$.}
\label{fig:method}
\end{figure*}


\section{Related Works}
\label{sec:related-works}
In this section, we examine related work from three distinct perspectives: (i) prompting methods that enhance CoTs through diversity, (ii) research focused on instruction tuning models using CoTs, and (iii) studies on self-correction in LLMs.

\paragraph{Improving Prompting through diversity.}
Many works have shown the benefits of generating diverse CoTs and aggregating them~\citep{wang2023selfconsistency, zhang2024self, yoran-etal-2023-answering, li2022making, weng-etal-2023-large, zhao-etal-2023-automatic, zhao-etal-2023-verify}. In particular, \citet{wang2023selfconsistency} proposed the creation of \textit{self-assembles} of CoTs to improve LLM's performance, which they call self-consistency. They sample a series of CoTs, select the most consistent answer, and show large performance gains on reasoning tasks. \citet{yoran-etal-2023-answering} extends this work by creating a meta prompt that aggregates the reasoning paths instead of selecting the most common answer. \citet{zhang2024self} propose explicit steps to contrast each CoT and reflect on the final answer. However, none of these works induce LLMs to generate multiple CoTs in the same inference step. 

\paragraph{Fine-Tuning on Diverse CoTs.} The success of CoT prompting led to the creation of instruction-tuning datasets with CoTs \citep{chung2024scaling}. \citet{kim-etal-2023-cot} argue that small LMs perform poorly on CoT on unseen tasks compared to large LMs. Hence, they create an instruction-tuning dataset of CoT to equip small LMs with CoT capabilities. Others suggest distilling CoTs from very large language models (vLLMs) \citep{hsieh-etal-2023-distilling, li-etal-2023-symbolic}. \citet{ho-etal-2023-large} also show the benefits of distilling CoTs from these vLLMs and claim that sampling multiple CoTs per question and training on these diverse CoTs is an effective data augmentation technique that improves the performance of distilled models. However, they do not use this diversity at inference time, and unlike us, their method only generates one CoT per question. \citet{ranaldi-freitas-2024-self} proposes a second step where the distilled student models generate multiple CoTs and with reinforcement learning the student trains itself. \citet{huang-etal-2023-large} show that vLLMs can improve performance on reasoning tasks by self-training on their own CoT generations from sampling. 

\paragraph{Self-Correction.} \citet{huang-etal-2023-large} defines it as the ability of an LLM to correct its initial response without relying on external feedback. Most works approach self-correction in LLMs with a system with two steps: one that generates the answer and another one that identifies errors \citep{shinn2024reflexion, madaan2023selfrefine, pan2023automatically, kim2024language, weng-etal-2023-large, jiang2023forwardbackward, multiagent_debate, paul-etal-2024-refiner, saunders2022self, akyurek-etal-2023-rl4f, welleck2023generating, estornell2024accdebate}. However, \citet{hong-etal-2024-closer} claims that LLMs cannot identify their own errors and \citet{huang2023large, stechly2024selfverification, tyen2023llms} argue that the self-corrections gains stem from the use of external feedback.
Our method differs from these methods in that we generate multiple CoTs in a single-inference step. As we will demonstrate, this access to previous CoTs, enables the model to refine subsequent reasoning chains without explicitly identifying errors. 

\section{Methods}

\paragraph{DCoT.} 
We instruction-tune LLMs on diverse CoTs to generate multiple CoTs in a single inference step.
To this end, we devise a DCoT instruction template, where we introduce a set of commands (in brackets) to request the number of CoTs to generate:

\begin{algorithmic}
\State \textbf{Prompt}: \texttt{[Question]} Question \texttt{[Options]} Options \texttt{[Number of answers]} $k$
\State \textbf{Response}: \texttt{[Answer 1]} CoT$_1$ \texttt{[Answer 2]} ... \texttt{[Answer k]} CoT$_k$ \texttt{[Final answer]} answer
\end{algorithmic}

In the input prompt, the instruction \texttt{[Options]} provides the candidate answers for multiple-choice questions answering tasks. For other tasks, such as span extraction, this is omitted.
In the response, the \texttt{[Final answer]} instruction is the convergence mechanism that conditions the model to generate the final answer.
We generate DCoT data in the required format using methods described in Section \ref{subsection:datasetgen}. 
For brevity, we refer to instruction-tuned models on DCoT data as \texttt{DCoT}.

\paragraph{CoT (Baseline).} So as to establish a comparable baseline, we instruction-tune the same LLMs using the more traditional CoT format. To ensure a fair comparison, we use the same reasoning chains as in \texttt{DCoT}. As shown in \Cref{fig:method}, each data point is composed of a question and a CoT, and a question may appear in more than one data point but with a different CoT. In this way, the model leverages CoT diversity at training time, but, unlike in \texttt{DCoT}, it does not do so at inference time. For brevity, we refer to these models as \texttt{CoT}.

With these two methods, we aim to compare two training regimes that use the same amount of training CoTs and where the only difference lies in the response format. We also do an exploratory analysis of whether we can replicate the results of our DCoT training with in-context learning in very large commercial language models in \Cref{app:prompting}.

\subsection{Training Data Generation}
\label{subsection:datasetgen}
We follow the methods set out by \citet{ott2023thoughtsource} to create CoTs for our CoT and DCoT datasets. We use GPT 3.5 turbo in the zero-shot setting with multiple triggers to generate CoTs. Specifically, \textit{CoT Triggers} are prompt suffixes, such as \textit{``Let's think step by step''} that `trigger' LLMs to generate CoTs. We use the same triggers as in \citep{ott2023thoughtsource}. For each question, we select four random CoT triggers. We limit the number of CoTs to four to ensure that the targets fit the context window of the LLMs. We restrict the training data to those reasoning chains that lead to correct answers as determined by the labels provided by the corresponding dataset\footnote{Initial experiments including wrong CoTs led to worse results, probably because in a supervised fine-tuning scenario they confused the model, so we only keep different CoTs that lead to correct answers.}. We report the prompt templates and triggers in \Cref{app:data_generation}.

\label{sec:datasets}
\Cref{tab:datasets_appendix} in \Cref{app:datasets} lists the datasets we use to generate our CoTs and train the models. These datasets were selected following prior works \citep{wang2023selfconsistency, yoran-etal-2023-answering}. We have added BoardgameQA \citep{kazemi2023boardgameqa} to include logic and ConditionalQA \citep{sun-etal-2022-conditionalqa} to include natural conditional reasoning, which is highly complex, and hence a revision of the answer can be beneficial. With this selection, we cover multiple domains, output spaces, and reasoning abilities. Following prior works \citep{khashabi-etal-2020-unifiedqa, pmlr-v202-longpre23a, wei2022finetuned, tafjord2021general}, we train all our models in all datasets at the same time to aim for generalizability. We provide more details on \Cref{app:datasets}.

\subsection{Models}
\label{subsection:models}
We train a series of models covering the scaling laws and different families. Concretely, we employ Phi 1.5 (1.3B; \citealt{li2023textbooks}), Phi 2 (2.7B; \citealt{phi2}), LLaMA-2 7B, LLaMA-2 13B \citep{touvron2023llama}. With this selection, we can analyze two families and scaling laws within the families. For all of our experiments, we select the non-instruction tuned-based models so as to ensure that the comparison between \texttt{DCoT} and \texttt{CoT} is fair. This is because instruction-tuning datasets contain CoT data \citep{touvron2023llama}, which would make the CoT baseline trained on longer and more diverse CoTs, and hence, the comparison between the two training regimes could be unfair. We also conduct a smaller experiment on LLaMA-2 13B Chat to analyze whether our \texttt{DCoT} instruction-tuning method can be applied to already-instruction-tuned models. Lastly, we also run our main experiments on LLaMA-2 70B. However, due to the inference costs, we train it on less data than the other models and only evaluate it on subsets of the evaluation set to show a hint of effectiveness on very large LMs.
We refer the reader to \Cref{app:hyperparamters} for details on the training setup of the models.

\subsection{Evaluation}
\label{sec:evaluation}
Our method is likely to be most effective in scenarios where access to previous CoTs and the corresponding answer is helpful, namely those tasks with a large output space. Therefore, to rigorously evaluate this, we test our method across tasks with varying output space sizes. Specifically, we assess our models on the following task types: Numeric, Span Extraction, Multiple-Choice, Binary, and Symbolic. We use the macro average F1 metric for all these in-domain classification tasks and the squad-metric \citep{rajpurkar-etal-2016-squad} for these in-domain span-extraction tasks (i.e., ConditionalQA and HotpotQA).
So as to select the value of the hyperparameter $k$, we run our \texttt{DCoT} with $k \in [1,4]$ and select the best $k$ for each dataset based on the dev set (\Cref{tab:bestk} in \Cref{app:bestk} reports them). All results are reported on the test set, with the exception of LLaMA02 70B. For LLaMA-2 70B, we only train on a small subset of our training set and also report results on subsets of the dev set with no hyperparameter optimization at all using a $k$ of 2 (the minimum amount of refinements), due to the costs. Further discussions are provided in \Cref{app:hyperparamters}.

\subsubsection{Unseen Tasks}
Generalization to new tasks remains a challenging problem. For example, \citet{chung2024scaling} and \citet{kim-etal-2023-cot} shows the need to train on thousands of tasks to achieve performance gains on unseen tasks with CoTs.
Since we advocate for the augmentation of instruction-tuning data with our DCoT, we need to evaluate that DCoT at least does not cause performance degradation on unseen tasks (i.e., tasks not used for training). Therefore, we select four challenging unseen tasks encompassing commonsense reasoning (CSQA; \citealt{talmor-etal-2019-commonsenseqa}), multiple-choice math (AQuA, \citealt{ling-etal-2017-program}), number generation for math (SVAMP, \citealt{patel-etal-2021-nlp}), and number generation for object counting (\citealt{suzgun2022challenging}).

\begin{table*}[t]
\centering
\scalebox{0.92}{
\setlength{\tabcolsep}{2pt}
\begin{tabular}{@{}llc|c|ccc|cccc|c||c@{}}
\toprule
             &  &  & \textbf{Numeric}                           & \multicolumn{3}{|c}{\textbf{Span-Extraction}}  & \multicolumn{4}{|c|}{\textbf{Multiple-Choice}}       &  \textbf{Binary} & \textbf{Symbolic} \\ \midrule
\textbf{LLM} & \textbf{Method} & \textbf{Avg.}  & \textbf{GSM8K}   & \textbf{CQA}             & \textbf{HQA}   & \textbf{Avg.}  & \textbf{ARC}             & \textbf{BGQA}  & \textbf{Quartz} & \textbf{Avg.}  & \textbf{StrQA}  & \textbf{LLC\textsuperscript{\textdagger}}  \\ \midrule
             & CoT             & 47.2           & 34.95            & 61.21                    & 32.56          & 46.88          & 48.7                     & 32.39          & 72.69           & 51.26          & 54.08           & \textbf{41}                                                               \\
Phi 1.5      & DCoT (Ours)     & \textbf{49.39} & \textbf{36.85}   & \textbf{62.48}           & \textbf{34.81} & \textbf{48.64} & \textbf{50.01}           & \textbf{38.6}  & \textbf{77.39}  & \textbf{55.34} & \textbf{55.97}  & 39                                                                        \\ \cline{2-13} 
(1.3B)       & CoT + SC        & 46.48          & \textbf{40.33}   & 63.39                    & 33.63          & 48.51          & \textbf{53.81}           & 21.59          & 75.11           & 50.17          & 51.96           & 32                                                                        \\
             & DCoT + SC       & \textbf{49.01} & 40.18            & \textbf{65.23}           & \textbf{37.79} & \textbf{51.51} & 53.24                    & \textbf{27.6}  & \textbf{81.08}  & \textbf{53.97} & \textbf{55.97}  & 31                                                                        \\ \hline
             & CoT             & 60.85          & 56.71            & 65.13                    & 52.65          & 58.89          & 70.87                    & 39.48          & 82.91           & 64.42          & \textbf{61.06}  & 58                                                                        \\
Phi 2        & DCoT            & \textbf{62.6}  & \textbf{60.73}   & \textbf{68.61}           & \textbf{55.15} & \textbf{61.88} & \textbf{73.77}           & \textbf{47.07} & \textbf{83.16}  & \textbf{68.00} & 54.34           & 58                                                                        \\ \cline{2-13} 
(2.7B)       & CoT + SC        & 61.5           & 64.97            & 68.14                    & 55.82          & 61.98          & 74.36                    & 28.99          & 85.2            & 62.85          & \textbf{59.51}  & 55                                                                        \\
             & DCoT + SC       & \textbf{65.12} & \textbf{68.08}   & \textbf{70.53}           & \textbf{58.61} & \textbf{64.57} & \textbf{76.06}           & \textbf{44.16} & \textbf{86.09}  & \textbf{68.77} & 51.43           & \textbf{66}                                                               \\ \hline
             & CoT             & 58.97          & 28.51            & 65.73                    & 53.88          & 59.80          & 61.63                    & \textbf{43.13} & 79.32           & 61.36          & \textbf{64.59}  & 75                                                                        \\
LLaMA2       & DCoT            & \textbf{60.8}  & \textbf{29.57}   & \textbf{70.99}           & \textbf{56.26} & \textbf{63.62} & \textbf{62.7}            & 41.91          & \textbf{81.37}  & \textbf{61.99} & 61.64           & \textbf{82}                                                               \\ \cline{2-13} 
7B           & CoT + SC        & \textbf{62.9}  & 33.97            & 69.92                    & 57.05          & 63.48          & 65.98                    & \textbf{46.04} & 83.28           & \textbf{65.10} & \textbf{65.99}  & 81                                                                        \\
             & DCoT + SC       & 61.09          & \textbf{36.01}   & \textbf{71.36}           & \textbf{58.35} & \textbf{64.85} & \textbf{68.53}           & 28.2           & \textbf{84.05}  & 60.26          & 59.22           & \textbf{83}                                                               \\ \hline
             & CoT             & 64.39          & 42.53            & 70.25                    & 60.23          & 65.24          & \textbf{71.79}           & 42.63          & \textbf{84.82}  & 66.41          & 61.85           & \textbf{81}                                                               \\
LLaMA2       & DCoT            & \textbf{66.18} & \textbf{44.28}   & \textbf{71.56}           & \textbf{63.52} & \textbf{67.54} & 71.41                    & \textbf{50.21} & 83.29           & \textbf{68.30} & \textbf{65.16}  & 80                                                                        \\ \cline{2-13} 
13B          & CoT + SC        & 66.82          & 50.27            & \textbf{72.72}           & 62.34          & 67.53          & 74.82                    & 40.8           & \textbf{85.84}  & \textbf{67.15} & 67.75           & 80                                                                        \\
             & DCoT + SC       & \textbf{68.12} & \textbf{54.51}   & 72.61                    & \textbf{65.92} & \textbf{69.26} & \textbf{74.89}           & \textbf{41.27} & 85.07           & 67.08          & \textbf{64.65}  & \textbf{86}                                                               \\ \hline

LLaMA2       & CoT             & 66.96          & 56               & \textbf{73.59}           & \textbf{55.94} & \textbf{64.76} & 81.69                    & \textbf{44.34} & 81.99           & 69.34          & 66.15           & 76                                                                        \\
70B*         & DCoT            & \textbf{68.63} & \textbf{66}      & 69.57                    & 49.78          & 59.67          & \textbf{89.04}           & 38.3           & \textbf{85.99}  & \textbf{71.11} & \textbf{68.34}  & \textbf{82}                                                               \\ \hline
LLaMA2       & CoT             & \textbf{64.87} & 42.76            & \textbf{71.71}           & 60.83          & 66.27          & 70.43                    & \textbf{44.39} & 84.04           & \textbf{66.29} & 66.78           & \textbf{78}                                                               \\
13B Chat\textdaggerdbl     & DCoT            & 64.62          & \textbf{44.2}    & 71.59                    & \textbf{63.87} & \textbf{67.73} & \textbf{72.22}           & 40.94          & \textbf{85.43}  & 66.20          & \textbf{67.68}  & 71                                                                        \\ \bottomrule
\end{tabular}
}
\caption{Comparison of \texttt{DCoT} against \texttt{CoT} on the test sets. We note the larger improvements on tasks with a larger output space (numeric and span extraction) in line with our hypothesis on where our method is more effective. *70B results on the dev set. \textdaggerdbl CoT in chat models include CoT data not used in DCoT. \textdagger Results on LLC are not reliable due to the small training and test set sizes.}
\label{tab:main_results}
\end{table*} 

\subsubsection{Robustness Check}

Lastly, we use Big Bench Hard \citep{suzgun2022challenging} for a control experiment to evaluate if our method hinders reasoning on complex tasks wherein CoT is beneficial only in models that are significantly larger than the ones we test; in other words, not using CoT is better for small models. This implies that it is extremely difficult for small models to generate correct CoTs for these tasks, and therefore, generating more than one is even more difficult, so it is reasonable to question whether DCoT can reduce performance.

\section{Results and Analysis}
In this section, we discuss our results with the aim of answering the following questions:
\begin{enumerate}
\itemsep0em 
    \item \textbf{Primary:} Does fine-tuning on DCoT (\texttt{DCoT}) provide performance gains compared to fine-tuning on CoT (\texttt{CoT}) with a comparable number of reasoning chains? (\Cref{sec:main_resuls})
    \item Given the inference cost, is \texttt{DCoT} effective on a reasonably small number of reasoning chains ($k$)? (\Cref{sec:dcot_k})
    \item Can \texttt{DCoT} be safely included in instruction-tuning datasets without unforeseen consequences on unseen tasks? (\Cref{sec:OOD}) and (\Cref{sec:BBH}) 
    \item Can \texttt{DCoT} benefit from \texttt{CoT} extensions? (\Cref{sec:dcot_sc})
    \item Are the improvements obtained a result of within-inference refinement or a result of random perturbations? (\Cref{sec:human_analysis} and \ref{sec:human_analysis_3}) 
\end{enumerate}

\subsection{DCoT is Beneficial on In-Domain Tasks}
\label{sec:main_resuls}
Our main goal in this experiment is to compare two training regimes: CoT and DCoT. The first two rows of each model in \Cref{tab:main_results} compare \texttt{DCoT} with the \texttt{CoT} baseline using the greedy decoding.\footnote{CoinFlip results are omitted because all models achieve perfect scores.}
The first result we observe is that \texttt{DCoT}, on average, is superior across all our main models. Analyzing the results per task type, we also observe that it is particularly effective on multiple-choice tasks, where the Phi models achieve a performance boost of four points. We also observe clear gains on span extraction and numeric tasks. However, performance on binary and symbolic tasks presents a more mixed picture. The challenges with StrategyQA (binary) can be attributed to its constrained output space (i.e., ``yes'' or ``no''), which limits the scope for effective revisions. For symbolic tasks, we use the Last Letter Concatenation dataset, which includes only 300 training and 100 test instances and is our smallest dataset (\Cref{app:datasets}). This small size reduces the reliability of the results, preventing us from drawing definitive conclusions. The overall performance of \texttt{DCoT} across all models confirms the effectiveness of DCoT training.

We also conduct a smaller experiment on general instruction-tuned models (LLaMA2 13B chat). It is worth noting that comparing \texttt{CoT} with \texttt{DCoT} is not completely fair in this setting because this model has already been fine-tuned on CoTs \citep{touvron2023llama}; thus, the \texttt{CoT} training is larger and more diverse than the \texttt{DCoT} one, as discussed in \Cref{subsection:models}. Despite this, \Cref{tab:main_results} shows that \texttt{DCoT} outperforms \texttt{CoT} in more than half of the datasets. The results of our other smaller experiment on LLaMA 2 70B also hint at the effectiveness of \texttt{DCoT} on larger models.

\subsection{One Refinement achieve Gains} 
\label{sec:dcot_k}

\begin{table}[t]
\begin{center}
\scalebox{0.79}{
\setlength{\tabcolsep}{2pt}
\begin{tabular}{@{}lccccc@{}}
\toprule
\textbf{Method} & \textbf{Phi 1.5} & \textbf{Phi 2} & \textbf{LL. 7B} & \textbf{LL. 13B} \\ \toprule
\textbf{CoT}     &   47.51±1.77             &         63.51±.71        &       59.30±.54      &       65.41±.91                              \\ \midrule
\textbf{DCoT@1}                  & 47.87±1.71                           & 63.91±2.58                         & 61.28±.50                              & 65.80±.44                               \\ \midrule
\textbf{DCoT@2}                  & 48.63±.67$\uparrow$                  & 65.33±2.80$\uparrow$               & 62.46±.45$\uparrow$                    & 67.30±.49$\uparrow$                     \\ \midrule
\textbf{DCoT@3}                  & 48.96±.66                            & 65.30±1.72                         & 62.37±.23                              & 66.92±.59                               \\ \midrule
\textbf{DCoT@4}                  & 48.76±.33                            & 64.89±2.39                         & 62.42±.59                              & 66.70±.55    \\ \bottomrule
\end{tabular}
}
\end{center}
\caption{\texttt{DCoT} average performance across different number of CoTs per question on the dev sets.}
\label{tab:dcot_k}
\end{table}


In this experiment, we examine the number of chains\footnote{Thanks to our \texttt{DCoT} training, the model generates exactly $k$ CoTs.} required to achieve performance gains relative to $k=1$. \Cref{tab:dcot_k} presents the average performance across all datasets for varying values of $k$. Notably, on average, $k=2$ consistently enhances performance across all models, supporting our hypothesis that DCoT effectively refines the initial response. However, increasing $k$ to three or four does not yield further improvements compared to $k=2$, except on GSM8k (math), where increasing $k$ up to $3$ consistently improves performance (see \Cref{tab:dev_set} in \Cref{app:dev_set}). We also observe similar patterns in our smaller experiments on LLaMA 13B Chat and LLaMA 2 70B (also in \Cref{tab:dev_set}). These results suggest that, on average, a single refinement is sufficient to improve performance, making it efficient and cheap. They also align with prior works \cite{estornell2024accdebate}, where 3 and 4 turns do not always improve performance, as we will discuss in \Cref{sec:discussion}.

\paragraph{\texttt{DCoT@1} $\approx$ \texttt{CoT}.}
An important phenomenon we observe in \Cref{tab:dcot_k} is that the performance of \texttt{DCoT} when generating a single CoT (i.e., \texttt{DCoT@1}) is very similar to the \texttt{CoT} baseline, as expected. This result shows that our \texttt{DCoT} training does not interfere with the regular CoT generation. \textit{Therefore, \texttt{DCoT} is a safe replacement to \texttt{CoT} in regular instruction-tuning datasets.}

\begin{table}[t]
\begin{center}
\scalebox{0.87}{
\setlength{\tabcolsep}{2pt}
\begin{tabular}{@{}llcccc@{}}
\toprule
\textbf{LLM}             & \textbf{Method} & \textbf{CSQA} & \textbf{AQuA} & \textbf{ObjCnt} & \textbf{SVAMP} \\ \midrule
\multirow{5}{*}{Phi 1.5} & CoT             & 33.88         & 20.27         & 35.60           & 40.00          \\ \cmidrule(l){2-6} 
                         & DCoT@1          & 32.26         & 21.51         & 25.20           & 40.50          \\
                         & DCoT@2          & 34.23         & 17.31         & 27.60           & 30.00          \\
                         & DCoT@3          & 33.81         & 22.38         & 30.80           & 30.00          \\
                         & DCoT@4          & 34.73         & 22.06         & 30.00           & 31.50          \\ \midrule
\multirow{5}{*}{Phi 2}   & CoT             & 44.29         & 29.52         & 54.00           & 55.00          \\ \cmidrule(l){2-6} 
                         & DCoT@1          & 44.15         & 34.86         & 58.40           & 60.50          \\
                         & DCoT@2          & 44.13         & 34.09         & 56.40           & 60.50          \\
                         & DCoT@3          & 45.99         & 31.83         & 57.60           & 60.00          \\
                         & DCoT@4          & 45.43         & 34.73         & 56.40           & 59.50          \\ \midrule
                         & CoT             & 38.41         & 19.41         & 34.80           & 39.50          \\ \cmidrule(l){2-6} 
                         & DCoT@1          & 36.94         & 17.70         & 40.00           & 41.50          \\
LLaMA2                   & DCoT@2          & 40.79         & 17.27         & 39.60           & 43.00          \\
7B                       & DCoT@3          & 40.67         & 16.90         & 36.80           & 43.00          \\
                         & DCoT@4          & 40.43         & 17.21         & 37.20           & 39.00          \\ \midrule
                         & CoT             & 46.55         & 24.85         & 45.2            & 62.50          \\ \cmidrule(l){2-6} 
                         & DCoT@1          & 44.62         & 23.98         & 46.00           & 55.00          \\
LLaMA2                   & DCoT@2          & 45.48         & 22.42         & 47.60           & 53.50          \\
13B                      & DCoT@3          & 47.42         & 20.72         & 52.40           & 56.50          \\
                         & DCoT@4          & 46.45         & 23.13         & 54.00           & 53.50          \\ \bottomrule
 
\end{tabular}
}
\end{center}
\caption{\texttt{CoT} and \texttt{DCoT} across various $k$ performance on unseen tasks.}
\label{tab:OOD_results}
\end{table}

\subsection{No Degradation in Unseen Tasks}
\label{sec:OOD}
In this section, we evaluate the performance of DCoT on unseen tasks to ensure that there is no unexpected deterioration.
\Cref{tab:OOD_results} shows that \texttt{DCoT} remains effective in Commonsense QA (multiple-choice), where it shows the same trend as in-domain scenarios (i.e., increasing $k$ yields better answers than the first one). However, the models show limited effectiveness in the tasks that require numbers (i.e., AQuA, ObjCnt, and SVAMP). When one generation yields worse answers, the next one usually corrects it (e.g., AQuA on Phi 1.5, SVAMP on LLaMA 13B). These limited gains in unseen math tasks are in line with expectations, given that LLMs' performance on arithmetical reasoning in out-of-domain settings is known to be poor \citep{qian-etal-2023-limitations}, and generalization to unseen tasks requires thousands of tasks \citep{kim-etal-2023-cot}, while we only trained on nine. Overall, while our results show that DCoT presents only small improvements in some of the cases, we note that there is no major deterioration.

\subsection{DCoT is Robust on Tasks where CoT is Detrimental}
\label{sec:BBH}

\begin{table}[t]
\begin{center}
\begin{tabular}{@{}lcccc@{}}

\toprule
\textbf{Method} & \textbf{Phi 1.5} & \textbf{Phi 2} & \textbf{LL. 7B} & \textbf{LL. 13B} \\ \midrule
CoT    & 28.37   & 46.7  & 31.08      & 36.38       \\
DCoT@1 & 28.31   & 44.56 & 31.23      & 34.59       \\
DCoT@2 & 28.07   & 45.81 & 31.11      & 35.94       \\
DCoT@3 & 28.35   & 45.92 & 31.00         & 36.90        \\
DCoT@4 & 28.21   & 46.71 & 31.13      & 36.45       \\ \bottomrule
\end{tabular}
\end{center}
\caption{Results on Big Bench Hard. LL stands for LLaMA2.}
\label{tab:bbh_cot}
\end{table}
We analyze the performance of our method on Big Bench Hard, a benchmark where small models do not benefit from CoTs \citep{suzgun2022challenging} to ensure that our method does not lead to unexpected performance degradation over the \texttt{CoT} baseline. The results from \Cref{tab:bbh_cot} show that on these tasks, \texttt{DCoT} exhibits similar performance to \texttt{CoT}, thus demonstrating that \texttt{DCoT} does not lead to deterioration in challenging cases, where CoT might be detrimental. Moreover, we can observe some performance gains on Phi 2 and LLaMA-2 13B when increasing $k$, further showing the robustness of DCoT tuning. These experiments allow us to conclude that there is no risk in adding DCoT training data to instruction tuning datasets. 

\subsection{DCoT Benefits from CoT Extensions}
\label{sec:dcot_sc}
DCoT aims to be a new CoT mechanism for instruction-tuning LLMs. To achieve this, we need to confirm that it is a safe drop-in replacement and works with any CoT extension, such as self-consistency \citep{wang2023selfconsistency}, a different decoding strategy. This decoding method is an add-on that has been shown to increase the performance of \texttt{CoT} across a wide range of tasks by sampling multiple generations and aggregating them by a voting mechanism.

The last two rows of each model (i.e., \texttt{CoT+SC} and \texttt{DCoT+SC}) in \Cref{tab:main_results} compare our \texttt{DCoT} with the \texttt{CoT} baseline using the self-consistency decoding with four samples. We observe that our \texttt{DCoT} also benefits from this mechanism and keeps its performance gains over the \texttt{CoT} baseline, showing that our method can be a replacement for \texttt{CoT} in future instruction-tuning datasets. It is also worth noting that, on average, our \texttt{DCoT} with the greedy decoding even outperforms \texttt{CoT+SC} on the Phi models, showing its high performance.

\subsection{Manual Analysis: DCoT@2}
\label{sec:human_analysis}
We conduct a manual evaluation to verify our conclusions based on the quantitative results. Due to the costs of manual evaluations, we conduct this only on LLaMA 7B, the mid-size model of our selection, to have insights that apply to our smaller and larger models. Specifically, we aim to understand how generating a second CoT can improve performance. To this end, we select instances for every dataset where LLaMA 7B with \texttt{DCoT@1} outputs an incorrect answer while \texttt{DCoT@2} results in a correct answer. Then, we randomly sample five of those instances per dataset, except in HotpotQA and LLC, where only three instances fulfilled the criteria, resulting in a total of 31 samples.

We note that the first reasoning chain of \texttt{DCoT@2} might differ from that of \texttt{DCoT@1} because they are different runs. We find this to be the case in ten instances. This implies that in most cases, the first CoT is the same in both cases. Of these instances where the first reasoning chain is shared, we observe that the second CoT of \texttt{DCoT@2} exhibits a different reasoning pattern from the first in 15 cases. Therefore, a second, improved CoT allows the model to generate a correct answer when the first CoT results in an incorrect answer. We also observe three cases where the second CoT is the same as the first one but corrects the logical conclusion. In two cases, the second CoT corrects reasoning steps from the first one, and in one case, the error of the first CoT had a slight error in writing the final answer that is corrected with the second one. Overall, these observations confirm that \texttt{DCoT} \textit{revises} its first answer by generating a second CoT.
\Cref{table:human_analysis_example} in \Cref{app:human_analysis} shows a few examples. 

\subsection{Manual Analysis: DCoT@3}
\label{sec:human_analysis_3}



\paragraph{Answers Patterns.} To understand the behavior of \texttt{DCoT@3}, we use GPT4o-mini to extract the three answers given by the model. We focus on the multiple-choice datasets to facilitate the answer extraction by GPT4o-mini. \Cref{tab:dcot_3_analysis} in \Cref{app:dcot3_ans_patterns} shows that the most common pattern is that the first answer is correct, and the subsequent CoTs reproduce it. This is expected because CoT achieves competitive performance. Moreover, this further evidences that the model does not randomly generate CoTs but tries to refine prior answers. Therefore, if the first answer is correct, the model does need to change it. More interestingly, we find 12 cases where the answer from the third CoT (a wrong answer) is not selected and instead selects prior answers (correct answers). In contrast, we also find 12 cases where the third CoT yields a new answer (the correct one), and the model selects it. These observations suggest that the model is not biased towards the last answer. We also observe 27 cases where a third wrong CoT changes a correct second CoT. However, we find 45 cases where the third CoT corrects the second one. We also provide an analysis of the error cases in \Cref{app:human_analysis}.

\section{Discussion}
\label{sec:discussion}
\paragraph{Training Method.} \texttt{DCoT} and \texttt{CoT} are trained on exactly the same amount of CoTs and questions. While the \texttt{CoT} baseline uses data points in the form of [(q, cot$_1$), (q, cot$_2$), ...],  \texttt{DCoT} uses data points in the form of [(q, cot$_1$, cot$_2$, ...) , ...]. In other words, a simple re-organization of the training CoTs into the form of multiple CoTs per label has a major impact on the model's performance. The goal of this training method is not to consistently generate different reasoning chains because if the first one is correct, there is no need to change it. This training method aims to enable LLMs to revise prior CoTs if needed.  Importantly, \texttt{DCoT@1} matches the performance of the \texttt{CoT} baseline, indicating that it is safe to augment existing instruction-tuning datasets with \texttt{DCoT} data, as it will not hinder model performance. Indeed, we envision a typical instruction-tuning dataset containing both DCoT and CoT data.

\paragraph{DCoT revises the first answer.}The results of our experiments indicate that \texttt{DCoT} is different from CoT-ensembling methods like self-consistency, which also benefit from generating multiple candidate answers but do so across different and independent inference steps using high-temperature values.
\texttt{DCoT}, instead, generates multiple CoTs in a single inference step, allowing each to be aware of the previous ones. As discussed in \Cref{sec:human_analysis} and \ref{sec:human_analysis_3}, this enables the model to revise earlier reasoning. Empirically, \texttt{DCoT@2} consistently outperforms \texttt{DCoT@1}, suggesting that such revision contributes to improved performance. However, \texttt{DCoT@3} and \texttt{DCoT@4} show diminishing returns, performing similarly to \texttt{DCoT@2}.
These findings show that after the first answer revision and associated performance gains, generating additional CoTs neither improves nor degrades performance, suggesting that the model cannot improve its answer further. This also confirms that the model does not simply generate a sequence of random CoTs but instead revises its first answer. The limitation on scaling $k$ aligns with observations from prior works on self-correction from \cite{madaan2023selfrefine, kim2024language}, where thanks to their external feedback (unlike us), they can achieve gains with up to $k=4$ and $k=3$ respectively. Furthermore, \citet{estornell2024accdebate} observe a similar behavior to our \texttt{DCoT}, most gains are obtained with $k=2$, and increasing $k$ further can lead to marginal gains or even small degradation.


\paragraph{Costs.}Lastly, the performance gains achieved by \texttt{DCoT} come at the cost of generating more tokens (i.e., a second CoT). This trade-off between performance and cost is common in prior works, such as in \citep{wang2023selfconsistency, yoran-etal-2023-answering, zhang2024self}. A CoT in our training dataset requires, on average, 124 Phi tokens and 142 LLaMA 2 tokens. Consequently, generating a second CoT to achieve performance gains would incur a negligible cost increment of \$0.00007 per model call,\footnote{Using the cost of LLaMA-2 13B on \href{https://replicate.com/meta/llama-2-13b}{replicate.com}
 (\$0.5/1M tokens).} making \texttt{DCoT} a cost-effective alternative to \texttt{CoT}.

\section{Conclusions}
This work presents Diverse Chain of Thought (\texttt{DCoT}), a new training method that aims to improve LLM's performance on reasoning tasks by generating multiple CoTs in a single inference step to revise answers. Through extensive quantitative experiments, we show the effectiveness and scalability of our method across a wide range of reasoning tasks, model families, and sizes. Furthermore, we show its efficiency by achieving gains with a single revision (i.e., $k=2$), which incurs negligible costs for the user. We also show that \texttt{DCoT} can be extended with any CoT extension, such as self-consistency, wherein it outperforms \texttt{CoT} similarly extended with self-consistency. Lastly, we show that the reason behind the performance gains of models fine-tuned with DCoT lies in its answer revising abilities, where a single answer revision is enough to achieve clear gains. We leave as future work extending our DCoT fine-tuning to other types of prompting methods such as code prompting \citep{puerto2024code} or graph of thoughts \citep{graph_of_thought}, and improving the scalability of $k$, without external feedback.

\section*{Limitations}
Our method is limited by the context window of the underlying model. In this work, we have explored generating CoTs up to $4$. Although we train the model to generate different chains, this does not guarantee it will generate intrinsically different chains at inference time, and this is not always desired (for example, once the model generates a correct answer, there is no need to change the reasoning chain). This is a limitation shared by prior works on generating diverse CoTs.

We limit the generation of the CoTs to a single commercial LLM provider because our preliminary experiments showed performance drops when combining multiple LLM providers. Further research on how to combine multiple LLM providers for distilling to smaller models is interesting and we leave that for future work.

Due to the computational costs, we could not extensively experiment on the 70B model. We could only afford to train with one seed and on a smaller dataset of 900 questions. Similarly, we could only evaluate it on 100 random questions per dataset. Nevertheless, the clear gains we observed on the dev sets, where we do not do any hyperparameter fine-tuning due to its costs, are indicative of the potential of our method on very large language models.

We do not conduct experiments on LLaMA 3 and 3.1 because, with them, we cannot do experiments across model sizes in a single model family as they only have 8B and 70B. LLaMA 3.2, on the other hand, was released after the conclusion of the core experiments and rerunning all experiments would not add new insights that justify the costs. Therefore, we conducted experiments with two families, one for smaller models (1.3B and 2.7B) and another family for larger but still manageable sizes (7B, 13B, and core experiments on 70B).

\section*{Ethics and Broader Impact Statement}
This work adheres to the ACL Code of Ethics. In particular, all the datasets we used have been shown by prior works to be safe for research purposes. They are not known to contain personal information or harmful content. Our method aims to improve the reasoning abilities of LLMs. Moreover, by generating multiple CoTs in one inference step, we allow the model to explore more reasoning chains and potentially diminish the effects of potentially biased or incorrect CoTs. Because of this, we believe our work can contribute to the safe deployment of LLMs in real-world scenarios.

\section*{Acknowledgments}
This research work has been funded by the German Research Foundation (DFG) as part of the UKP-SQuARE project (grant GU 798/29-1) and by the German Federal Ministry of Education and Research and the Hessian Ministry of Higher Education, Research, Science and the Arts within their joint support of the National Research Center for Applied Cybersecurity ATHENE. We would also like to thank the Early Career Research grant from the University of Bath.

We gratefully acknowledge the support of Microsoft with a grant for access to OpenAI GPT models via the Azure cloud (Accelerate Foundation Model Academic Research).

Lastly, we thank Irina Bigoulaeva, Sheng Lu, Subhabrata Dutta, and the anonymous reviewers for their insightful feedback on a prior version of this manuscript.

\bibliography{anthology,custom}

\appendix

\section{Datasets}
\label{app:datasets}
All the datasets used in this work are exclusively in English language. In particular, we use ARC \citep{clark2018think}, BGQA          \citep{kazemi2023boardgameqa},CoinFlip        \cite{NEURIPS2022_9d560961}, ConditionalQA (CQA)  \citep{sun-etal-2022-conditionalqa}, GSM8K        \citep{cobbe2021training}, HotpotQA  (HQA)   \citep{yang-etal-2018-hotpotqa}, LLC            \cite{NEURIPS2022_9d560961}, Quartz         \citep{tafjord-etal-2019-quartz}, and StrategyQA (StrQA)     \citep{geva-etal-2021-aristotle}  for training, while we use AQuA \citep{ling-etal-2017-program}, CommonsenseQA \citep{talmor-etal-2019-commonsenseqa}, Object Count (a task of Big Bench Hard \citealt{suzgun2022challenging}), SVAMP \citep{patel-etal-2021-nlp}, and Big Bench Hard for out of domain evaluation. For BGQA, we use the partition \texttt{main-3}, the most difficult one requiring 3-hop reasoning skills.

Some of these datasets do not provide a validation set. In those cases, we randomly sample 500 instances from the training set and use them as validation set. Similarly, when a dataset does not provide a test set, we use the validation set as a test and create a validation set from the unused instances from the training set. When the training set is not larger than 1k, we divide the validation set into two. For Last Letter Concatenation (LLC), the training set is very small (350 instances), and the test set is also very small (150), so we pick 50 instances of the test set as validation and 100 as test. We release in our github repository the exact partitions we used.

\Cref{tab:datasets_appendix} reports the licenses and sizes of the training, dev, and test sets of the datasets we used and \Cref{tab:ooddatasets_appendix} reports for the out of domain datasets. We use these datasets for research purposes only, fulfilling their intended use.

Due to the large size of LLaMA-2 70B and its computation costs, we trained it on a smaller sample data of 900 questions. Similarly, for inference, we pick a random sample of 100 questions per dataset.

\section{Experimental Setup}
\label{app:hyperparamters}
We run all our experiments on a GPU cluster with an Nvidia A180. To run GPT models, we use the Azure OpenAI service and prompt them with the library Langchain.\footnote{\url{https://github.com/langchain-ai/langchain}} We use Scikit-learn \citep{scikit-learn} for the implementation of the evaluation metrics.

We train all models using LoRA \citep{hu2022lora} with the PEFT library \citep{peft} and use vLLM \citep{kwon2023efficient} as the inference engine. For training, we load the models with fp8, while for inference, we load them with fp16. We train models for three epochs, save checkpoints for each epoch and select the best checkpoint based on the average results on the dev set. 

Due to the challenge of running very large models, such as LLaMA-2 70B, to simplify the evaluation setup. We trained the model with 8-bit quantization and ran the evaluation on 4-bit. Instead of evaluating on the full dev sets, we had to evaluate on a random sample of 100 questions per dataset and only evaluate the last checkpoint. Therefore, we could not conduct hyperparameter tuning either. Because of these challenges, we cannot report results on the test set, and instead, we only report results on the dev set. It is important to emphasize again that we do not conduct any hyperparameter tuning, so the results on the dev set are representative of the performance of our method on large-scale models.

\Cref{tab:training_time} shows the traning time needed for both methods across all our models. The parameters we use to train the models are reported in \Cref{tab:training_params}. \Cref{tab:hyperparameters} reports the best hyperparameters we found on the dev set.

\begin{table}[]
\begin{tabular}{@{}lllll@{}}
\toprule
\textbf{Method} & \textbf{Phi 1.5} & \textbf{Phi 2} & \textbf{LL. 7B} & \textbf{LL. 13B} \\ \midrule
DCoT   & 12h     & 20h   & 35h      & 51h       \\
CoT    & 9h      & 15h   & 26h      & 39h       \\ \bottomrule
\end{tabular}
\caption{Training time for \texttt{DCoT} and \texttt{CoT}}
\label{tab:training_time}
\end{table}

\begin{table}[t]
\begin{center}
\begin{tabular}{@{}lc@{}}
\toprule
\textbf{Param. name} & \textbf{Value}\\
\midrule
lora\_r                         & 64        \\ 
lora\_alpha                     & 16                         \\
lora\_dropout                   & 0.1                        \\
batch size & 4 \\
max\_grad\_norm                 & 0.3                        \\
learning\_rate                  & 2e-4                       \\
weight\_decay                   & 0.001                      \\
optim                           & paged\_adamw\_32bit        \\
lr\_scheduler\_type             & constant                   \\
max\_steps                      & -1                         \\
warmup\_ratio                   & 0.03                       \\
group\_by\_length               & True                       \\
max\_seq\_length                & 4096                       \\
packing                         & False                      \\
seeds & 0, 42, 2024\\
load\_in\_8bit & True\\
\bottomrule
\end{tabular}
\caption{Training parameters}
\label{tab:training_params}
\end{center}
\end{table}
\begin{table}[t]
\begin{center}
\begin{tabular}{@{}lccc@{}}
\toprule
\textbf{Model}              & \textbf{Method} & \textbf{Seed} & \textbf{Epoch} \\ \midrule
\multirow{2}{*}{Phi 1.5}    & CoT             & 0             & 2              \\
                            & DCoT            & 42            & 2              \\
\multirow{2}{*}{Phi 2}      & CoT             & 0             & 3              \\
                            & DCoT            & 2024          & 2              \\
\multirow{2}{*}{LLaMA2 7B}  & CoT             & 0             & 2              \\
                            & DCoT            & 0             & 3              \\
\multirow{2}{*}{LLaMA2 13B} & CoT             & 42            & 3              \\
                            & DCoT            & 42            & 3              \\ \bottomrule
\end{tabular}
\caption{Best hyperparameters tuned on the dev set.}
\label{tab:hyperparameters}
\end{center}
\end{table}

\begin{table*}[t]
\begin{center}
\setlength{\tabcolsep}{2pt}
\begin{tabular}{lp{2cm}cccccp{2cm}c}
\toprule
\textbf{Dataset}  & \textbf{Task}     & \textbf{Train} & \textbf{Dev} & \textbf{Test} & \textbf{License}  & \textbf{Source}  & \textbf{Description} & \textbf{Citation}     \\ \midrule
ARC     & Multiple choice      & 1033  & 294 & 1150 &   CC BY-SA 4.0 & \href{https://huggingface.co/datasets/allenai/ai2_arc}{Link} & Science & \citep{clark2018think}  \\
BGQA    & Multiple choice      & 716  & 500 &   1000 & CC BY & \href{https://storage.googleapis.com/gresearch/BoardgameQA/BoardgameQA.zip}{Link}         & Logic & \citep{kazemi2023boardgameqa} \\
Coin Flip  & Multiple choice    &  1000 &  1333  & 3333 & mit & \href{https://huggingface.co/datasets/skrishna/coin_flip}{Link}  & State-tracking & \citep{NEURIPS2022_9d560961} \\
CQA       & Span extraction     & 958 & 285 &  804 &  CC BY-SA 4.0 & \href{https://haitian-sun.github.io/conditionalqa/}{Link} & Conditional & \citep{sun-etal-2022-conditionalqa})   \\
GSM8K   & Num. Generation      &    1000   & 500 & 1319 &  mit  & \href{https://huggingface.co/datasets/openai/gsm8k}{Link}    & Math & \citep{cobbe2021training}    \\
HQA  & Span extraction   & 1000 & 500 &  7405 & CC BY-SA 4.0 & \href{https://hotpotqa.github.io/}{Link}  & Explicit multi-hop & \citep{yang-etal-2018-hotpotqa})    \\
LLC    & Generation       & 350 & 50 & 100 & N/A & \href{https://huggingface.co/datasets/ChilleD/LastLetterConcat}{Link}   & Symbolic     & \citep{NEURIPS2022_9d560961}\\
Quartz   & Multiple choice     & 953 & 384 & 784 & CC BY-SA 4.0 & \href{https://huggingface.co/datasets/allenai/quartz}{Link}    & Relationships & \citep{tafjord-etal-2019-quartz} \\
StrQA & Boolean QA & 998 & 343 & 344 & mit & \href{https://huggingface.co/datasets/ChilleD/StrategyQA}{Link}   & Implicit multi-hop & \citep{geva-etal-2021-aristotle}) \\ \bottomrule
\end{tabular}
\end{center}
\caption{Training datasets. The training size corresponds to our CoT generations to build the DCoT dataset.}
\label{tab:datasets_appendix}
\end{table*}

\begin{table*}[]
\begin{center}
\begin{tabular}{lp{3cm}cccc}
\toprule
\textbf{Dataset}    & \textbf{Task}  &  \textbf{Dev}  & \textbf{License}  & \textbf{Source} & \textbf{Citation}       \\ \midrule
AQuA    & Multiple choice Math       &  254 &   Apache 2.0 & \href{https://github.com/google-deepmind/AQuA}{Link} & \citep{ling-etal-2017-program} \\
CSQA    & Multiple choice commonsense     &    1220     & mit & \href{https://huggingface.co/datasets/tau/commonsense_qa/}{Link} & \citep{talmor-etal-2019-commonsenseqa} \\
SVAMP  & Generation (numbers)    &   100   & mit & \href{https://github.com/arkilpatel/SVAMP}{Link} & \citep{patel-etal-2021-nlp} \\
Big Bench Hard   & Multiple choice \& Generation         &   6511 &  mit & \href{https://huggingface.co/datasets/maveriq/bigbenchhard}{Link} & \citep{suzgun2022challenging}  \\
 \bottomrule
\end{tabular}
\caption{\label{tab:ooddatasets_appendix} Out of domain datasets.}
\end{center}

\end{table*}

\begin{table}[]
\centering
\begin{tabular}{@{}lccc@{}}
\toprule
\textbf{Pattern}                 & \textbf{ARC} & \textbf{BGQA} & \textbf{Quartz} \\ \midrule
AAA $\rightarrow$ A     & 226  & 421    & 320 \\
AAB $\rightarrow$ A (o) & 6    & 4      & 2   \\
AAB $\rightarrow$ A (x) & 2    & 2      & 1   \\
AAB $\rightarrow$ B (o) & 2    & 3      & 7   \\
AAB $\rightarrow$ B (x) & 6    & 2      & 5   \\
ABA $\rightarrow$ A (o) & 4    & 8      & 12  \\
ABA $\rightarrow$ A (x) & 5    & 1      & 1   \\
ABA $\rightarrow$ B (o) & 0    & 0      & 0   \\
ABA $\rightarrow$ B (x) & 0    & 0      & 0   \\
ABB $\rightarrow$ A (o) & 0    & 2      & 2   \\
ABB $\rightarrow$ A (x) & 2    & 2      & 2   \\
ABB $\rightarrow$ B (o) & 10   & 8      & 16  \\
ABB $\rightarrow$ B (x) & 17   & 22     & 11  \\
ABC $\rightarrow$ A     & 1    & 0      & 1   \\
ABC $\rightarrow$ B     & 2    & 0      & 0   \\
ABC $\rightarrow$ C     & 2    & 1      & 0  \\\bottomrule
\end{tabular}
\caption{Answer pattern in \texttt{DCoT@3} for multiple choice datasets. (o) means correct final answer and (x) incorrect.}
\label{tab:dcot_3_analysis}
\end{table}

\section{Prompting}
\label{app:prompting}
We conducted exploratory experiments to evaluate the effectiveness of DCoT prompting on commercial black-box LLMs. 
We use prompts to require models to generate multiple CoTs, compare them, and generate an answer, all in a single inference step. We found that smaller LLMs, with fewer than 100B  parameters, lacked the capacity to perform this complex task. When prompted, they often generated the same CoT repeatedly. Even when they did generate multiple CoTs, our manual evaluation revealed they failed to effectively select the correct answer from among them. These results are in line with prior results that indicate that these smaller models are also not the most effective in generating CoTs~\citep{kim-etal-2023-cot}. While GPT-4o showed more success, the complexity of the task also heightened its tendency to hallucinate.
Consequently, we observed no performance boost through prompting alone and thus focused our experiments on instruction tuning using DCoTs, as detailed in subsequent sections.

The prompts we used with GPT4o for DCoT and CoT are ``Generate {k} different reasoning chains that answer the question. Make sure that none of the reasoning chains are repeated. Generate each reasoning chain independently, and not based on previous reasoning chains. This means that each reasoning chain must be as different from the others as possible. When generating the different reasoning chains, do so without knowledge of the answer. Each step in each of the reasoning chains must build on the previous steps in that reasoning chain. Once the required number of reasoning chains are generated, generate an answer based on the all the answers generated by all the reasoning chains.'' and ``Generate a reasoning chain that answer the question.'' In both cases, after generating the CoT, we extracted the answer with the following prompt for SVAMP ``Therefore, based on the solution above, extract the number that represents the answer:'' and ``Therefore, based on the solution above, select one of the options ({options}) as the answer to the question (just give me the option and nothing else).'' for ARC and Quartz.

\section{Dev Set Results}
\label{app:dev_set}
We report the mean and stardard deviation results from the validation set across threee random seeds in \Cref{tab:dev_set}.
\begin{table*}[t]
\begin{center}
\scalebox{0.73}{
\renewcommand{\arraystretch}{0.85}
\setlength{\tabcolsep}{2.pt}
\begin{tabular}{@{}lllccccccccc@{}}
\toprule
\textbf{LLM} &
  \textbf{Method} &
  \textbf{k} &
  \textbf{Avg} &
  \textbf{ARC} &
  \textbf{BGQA} &
  \textbf{CQA} &
  \textbf{GSM8K} &
  \textbf{HQA} &
  \textbf{LLC} &
  \textbf{Quartz} &
  \textbf{StrQA} \\ \midrule
\multirow{5}{*}{Phi 1.5} & \multirow{4}{*}{DCoT} & 1 & 47.87±1.71 & 44.13±1.94 & 39.43±3.91 & 61.83±.74 & 36.07±1.70 & 38.70±3.18 & 36.00±3.46 & 71.69±1.73 & 55.13±.35 \\
 &
   &
  2 &
  48.63±0.67 &
  46.98±2.60 &
  41.94±3.10 &
  60.87±1.14 &
  38.80±3.10 &
  39.79±3.80 &
  30.00±4.00 &
  74.29±2.69 &
  56.40±.87 \\
 &
   &
  3 &
  \textbf{48.96±0.66} &
  47.32±1.66 &
  42.75±1.92 &
  60.75±1.15 &
  39.00±1.71 &
  38.19±2.81 &
  32.67±7.02 &
  75.42±2.38 &
  55.57±1.52 \\
 &
   &
  4 &
  48.76±0.33 &
  46.78±1.14 &
  43.23±2.22 &
  60.16±1.32 &
  38.93±3.31 &
  37.33±2.92 &
  32.67±7.02 &
  75.60±3.32 &
  55.41±1.30 \\
 &
  CoT &
   &
  47.51±1.77 &
  46.60±2.38 &
  36.65±1.90 &
  59.55±0.61 &
  37.40±3.22 &
  35.28±4.22 &
  36.67±9.02 &
  75.07±2.36 &
  52.84±2.47 \\ \midrule
\multirow{5}{*}{Phi 2}   & \multirow{4}{*}{DCoT} & 1 & 63.91±2.58 & 75.21±1.84 & 45.01±3.03 & 65.39±1.57 & 56.47±1.68 & 62.44±2.63 & 62.67±16.29 & 82.88±1.09 & 57.28±2.35 \\
 &
   &
  2 &
  \textbf{65.33±2.80} &
  76.46±2.52 &
  46.89±3.85 &
  65.69±2.12 &
  57.60±1.64 &
  63.71±2.18 &
  66.67±9.02 &
  84.10±1.36 &
  56.44±3.33 \\
 &
   &
  3 &
  65.30±1.72 &
  75.87±1.42 &
  48.06±1.75 &
  65.90±2.02 &
  58.20±1.91 &
  61.66±2.06 &
  68.00±5.29 &
  83.91±1.18 &
  56.28±3.90 \\
 &
   &
  4 &
  64.89±2.39 &
  75.91±2.72 &
  49.11±2.31 &
  65.92±1.01 &
  57.07±1.33 &
  59.86±.96 &
  66.00±8.00 &
  84.09±1.88 &
  56.97±5.00 \\
 &
  CoT &
   &
  63.51±.71 &
  74.19±.61 &
  42.08±.79 &
  66.92±.29 &
  62.80±3.53 &
  56.45±.78 &
  62.71±3.00 &
  77.92±7.30 &
  66.74±15.54 \\ \midrule
LLaMA-2 &
  \multirow{4}{*}{DCoT} & 1 &
   61.28±.50 & 59.36±2.29 & 43.67±.35 & 65.31±.50     & 29.73±1.63 & 62.92±3.16 & 86.67±2.31 & 80.63±.92  & 61.96±1.45    \\
 7B&
   &2
   &
   \textbf{62.46±.45} & 61.63±1.46 & 43.56±.80 & 66.05±.80     & 33.40±.80  & 63.86±1.23 & 86.67±3.06 & 82.11±1.57 & 62.38±1.21   \\
 &
   &3
   &
   62.37±.23 & 60.98±2.37 & 44.23±.95 & 66.65±1.21    & 33.53±.50  & 63.46±1.46 & 86.67±1.15 & 80.89±2.65 & 62.51±.86  \\
 &
   &4
   &
  62.42±.59  & 62.13±3.21 & 43.85±.45  & 65.98±2.72 & 33.33±.50  & 63.63±2.16 & 86.00±3.46 & 82.20±2.78 & 62.20±1.42 \\
 &
  CoT & & 59.30±.54 & 56.54±3.83 & 41.91±2.32 & 59.85±3.91    & 31.93±1.42 & 57.81±3.73 & 82.67±3.06 & 79.24±2.16 & 64.42±1.52
   \\ \midrule
LLaMA-2 &
  \multirow{4}{*}{DCoT} & 1
   &65.80±.44 & 71.08±1.95 & 46.92±3.13 & 67.97±1.48 & 42.40±.35  & 69.97±.61  & 76.00±5.29  & 85.07±.40 & 66.99±2.10
   \\
 13B& & 2 
   &
   \textbf{67.30±.49} & 74.85±1.68 & 46.40±4.13 & 68.55±1.33 & 44.53±1.51 & 72.35±.93  & 81.33±3.06  & 84.89±.90 & 65.46±1.17
   \\
 &
   & 3 
   &
   66.92±.59 & 73.29±2.17 & 48.65±4.48 & 67.32±2.19 & 46.53±1.89 & 71.17±2.07 & 74.00±15.62 & 84.89±.52 & 63.07±.49
   \\
 &
   & 4
   &66.70±.55 & 74.17±2.78 & 46.58±3.03 & 68.46±1.40 & 43.93±3.42 & 70.92±3.54 & 78.67±3.06  & 82.46±.99 & 64.23±2.56
   \\
 &
  CoT & 
   &65.41±.91 & 71.66±2.15 & 44.45±1.53 & 68.39±1.70 & 42.67±2.32 & 66.12±.82  & 82.00±5.29  & 82.37±.82 & 65.64±1.29
   \\ \midrule
LLaMA-2 &
  \multirow{4}{*}{DCoT} & 1 &
   64.53 &	71.85	 &47.11 &	67.37 &	41.60 &	70.52 &	68.00 &	82.81 &	66.97
   \\
13B Chat* & & 2
   &
   65.95&	70.73&	47.76&	69.16&	42.40&	71.02&	74.00&	84.87&	67.68
   \\
 &
   & 3 &
   66.10&	72.22&	46.82&	67.48&	43.60&	72.08&	76.00&	84.87&	65.76
   \\
 &
   & 4
   &66.17&	71.85&	45.03&	69.70&	45.00&	71.75&	74.00&	86.44&	65.59
   \\
 &
  CoT      & & \textbf{66.27}   & 70.43 & 45.36 & 70.71         & 44.20 & 70.11    & 80.00 & 82.53  & 66.78      \\
\midrule
LLaMA-2 &
  \multirow{4}{*}{DCoT} & 1 &
   66.48&	85.80&	36.92&	65.89&	56.00&	49.78&	78.00&	87.00&	72.41
   \\
70B* & & 2
   &
   \textbf{68.63}	& 89.04	&38.30&	69.57&	66.00 &	49.78&	82.00&	85.99&	68.34
   \\
 &
   & 3 &
   68.07	&86.64&	38.55&	68.87&	60.00&	49.78&	80.00&	88.00&	72.70
   \\
 &
   & 4
   &68.06&	88.94&	32.62&	67.21&	66.00&	49.78&	82.00&	85.00&	72.92
   \\
 &
  CoT      & & 66.96&	81.69&	44.34&	73.59&	56.00&	55.94&	76.00&	81.99&	66.15     \\
                                \bottomrule
\end{tabular}
}
\caption{Dev set results using 3 random seeds. *One seed for LLaMA-2 13B Chat and 70B.}
\label{tab:dev_set}
\end{center}
\end{table*}


\section{DCoT across $k$ per Task Type}
\label{app:dcot_k}
\Cref{tab:dcot_k_per_task_type} shows the performance of \texttt{DCoT} across $k$ per task type. We observe that 2 revisions (i.e., $k=3$) is beneficial for math reasoning, while multiple choice and span extraction only needs one revision. As we mentioned in the main paper, symbolic results are not reliable due to the small training and set sizes of the dataset.
\begin{table*}[]
\centering
\scalebox{0.75}{
\setlength{\tabcolsep}{2.pt}
\begin{tabular}{@{}llcccccc@{}}
\toprule
\textbf{Model}             & \textbf{k} & \textbf{MC Avg.} & \textbf{Span Avg.} & \textbf{Math} & \textbf{Binary} & \textbf{Symbolic*} & \textbf{Total Avg} \\ \midrule
\multirow{5}{*}{Phi 1.5}   & CoT        & 52.77±1.55  & 47.41±2.41    & 37.40±3.22    & 52.84±2.47      & 36.67±9.02        & 47.51±1.77         \\\cmidrule(l){2-8} 
                           & 1          & 51.75±2.17  & 50.26±1.71    & 36.07±1.70    & 55.13±.35       & 36.00±3.46        & 47.87±1.71         \\
                           & 2          & 54.40±1.52  & 50.33±2.25    & 38.80±3.10    & 56.40±.87       & 30.00±4.00        & 48.63±.67          \\
                           & 3          & 55.16±1.11  & 49.47±1.98    & 39.00±1.71    & 55.57±1.52      & 32.67±7.02        & 48.96±.66          \\
                           & 4          & 55.20±.88   & 48.74±2.05    & 38.93±3.31    & 55.41±1.30      & 32.67±7.02        & 48.76±.33          \\ \midrule
\multirow{5}{*}{Phi 2}     & CoT        & 65.75±.33   & 62.26±.78     & 62.80±3.53    & 66.74±15.54     & 62.71±3.00        & 63.51±.71          \\\cmidrule(l){2-8} 
                           & 1          & 67.70±1.74  & 63.92±1.36    & 56.47±1.68    & 57.28±2.35      & 62.67±16.29       & 63.91±2.58         \\
                           & 2          & 69.15±1.72  & 64.70±1.12    & 57.60±1.64    & 56.44±3.33      & 66.67±9.02        & 65.33±2.80         \\
                           & 3          & 69.28±.33   & 63.78±1.52    & 58.20±1.91    & 56.28±3.90      & 68.00±5.29        & 65.30±1.72         \\
                           & 4          & 69.70±1.33  & 62.89±.63     & 57.07±1.33    & 56.97±5.00      & 66.00±8.00        & 64.89±2.39         \\\midrule
\multirow{5}{*}{LLaMA 7B}  & CoT        & 57.09±.96   & 58.11±1.12    & 31.93±1.42    & 64.42±1.52      & 82.67±3.06        & 59.30±.54          \\\cmidrule(l){2-8} 
                           & 1          & 61.22±.92   & 64.12±1.74    & 29.73±1.63    & 61.96±1.45      & 86.67±2.31        & 61.28±.50          \\
                           & 2          & 62.43±.56   & 64.96±1.01    & 33.40±.80     & 62.38±1.21      & 86.67±3.06        & 62.46±.45          \\
                           & 3          & 62.03±.21   & 65.06±1.30    & 33.53±.50     & 62.51±.86       & 86.67±1.15        & 62.37±.23          \\
                           & 4          & 62.73±.38   & 64.81±2.34    & 33.33±.50     & 62.20±1.42      & 86.00±3.46        & 62.42±.59          \\\midrule
\multirow{5}{*}{LLaMA 13B} & CoT        & 65.80±1.02  & 65.66±2.28    & 42.67±2.32    & 65.64±1.29      & 82.00±5.29        & 65.41±.91          \\\cmidrule(l){2-8} 
                           & 1          & 67.69±1.20  & 68.97±.52     & 42.40±.35     & 66.99±2.10      & 76.00±5.29        & 65.80±.44          \\
                           & 2          & 68.72±1.49  & 70.45±1.13    & 44.53±1.51    & 65.46±1.17      & 81.33±3.06        & 67.30±.49          \\
                           & 3          & 67.91±1.25  & 69.85±1.30    & 46.53±1.89    & 63.07±.49       & 74.00±15.62       & 66.92±.59          \\
                           & 4          & 68.17±1.58  & 70.00±1.16    & 43.93±3.42    & 64.23±2.56      & 78.67±3.06        & 66.70±.55          \\ \bottomrule
\end{tabular}
}
\caption{\texttt{DCoT} performance across k for each task type. *Symbolic results are not reliable due to the small size of the dataset LLC.}
\label{tab:dcot_k_per_task_type}
\end{table*}

\section{OOD Results on Tasks requiring Number Generation}
\label{app:number_ood}
\Cref{tab:ood_numerical} shows the average results across all out-of-domain datasets that requires number generation for all tasks. We can see that the average performance across all numerical tasks remains relatively similar across all $k$ values for all models except Phi-1.5, the smallest and weakest model. 

\begin{table*}[]
\centering
\begin{tabular}{@{}llcccc@{}}
\toprule
\textbf{LLM}                & \textbf{Method} & \textbf{Average} & \textbf{AQuA} & \textbf{ObjCnt} & \textbf{SVAMP} \\ \midrule
\multirow{5}{*}{Phi 1.5}    & CoT             & 31.96            & 20.27         & 35.6            & 40             \\
                            & DCoT@1          & 29.07            & 21.51         & 25.2            & 40.5           \\
                            & DCoT@2          & 24.97            & 17.31         & 27.6            & 30             \\
                            & DCoT@3          & 27.73            & 22.38         & 30.8            & 30             \\
                            & DCoT@4          & 27.85            & 22.06         & 30              & 31.5           \\\midrule
\multirow{5}{*}{Phi 2}      & CoT             & 46.17            & 29.52         & 54              & 55             \\
                            & DCoT@1          & 51.25            & 34.86         & 58.4            & 60.5           \\
                            & DCoT@2          & 50.33            & 34.09         & 56.4            & 60.5           \\
                            & DCoT@3          & 49.81            & 31.83         & 57.6            & 60             \\
                            & DCoT@4          & 50.21            & 34.73         & 56.4            & 59.5           \\ \midrule
\multirow{5}{*}{LLaMA2 7B}  & CoT             & 31.24            & 19.41         & 34.8            & 39.5           \\
                            & DCoT@1          & 33.07            & 17.7          & 40              & 41.5           \\
                            & DCoT@2          & 33.29            & 17.27         & 39.6            & 43             \\
                            & DCoT@3          & 32.23            & 16.9          & 36.8            & 43             \\
                            & DCoT@4          & 31.14            & 17.21         & 37.2            & 39             \\ \midrule
\multirow{5}{*}{LLaMA2 13B} & CoT             & 44.18            & 24.85         & 45.2            & 62.5           \\
                            & DCoT@1          & 41.66            & 23.98         & 46              & 55             \\
                            & DCoT@2          & 41.17            & 22.42         & 47.6            & 53.5           \\
                            & DCoT@3          & 43.21            & 20.72         & 52.4            & 56.5           \\
                            & DCoT@4          & 43.54            & 23.13         & 54              & 53.5   \\ \bottomrule        
\end{tabular}
\caption{\texttt{DCoT} and \texttt{CoT} results on out-of-domain numerical reasoning tasks.}
\label{tab:ood_numerical}
\end{table*}

\section{Manual Analysis}
\label{app:human_analysis}
\Cref{app:human_analysis} shows two examples of how the second CoT of LLaMA 7B with DCoT corrects the first CoT.

\paragraph{Error Cases.}To understand why \texttt{DCoT@3} does not further improve performance on average compared to $k=2$, we randomly sample five instances per dataset where \texttt{DCoT@3} fails to give the right answer and analyze the reasoning chains. We observe that an overwhelming majority of 26/40 exhibit the same second and third CoT and, therefore, do not improve nor decrease performance. We also observe four cases where the second and third CoTs show very similar semantics but different wording, two cases where the first and third CoTs are the same, two cases where all the CoTs are the same, two cases where the CoTs are similar but the arithmetic results are different, three cases where all the CoTs are different, and once case where all the answers are correct but do not match the label.
This analysis shows that one of the reasons for the similar average performance of $k=2$ and $k=3$ is that \texttt{DCoT} struggles to make a third better answer. This suggests that the model is not generating a series of random CoTs but it is trying to revise its prior answers. This limitation shown in these small fine-tuned models is also seen on very large LMs with prompting \Cref{app:prompting}.

\begin{table*}
    
    \begin{center}
    \begin{tabular}{p{0.12\linewidth} | p{0.80\linewidth} }
    \toprule
        \textbf{Type} & \textbf{Text}  \\
        \midrule         
1st and 2nd CoTs are the same but the 1st CoT makes an error when generating the answer. 2nd CoT corrects that. 

        &  \textbf{Question}: Take the last letters of each words in ``Billie Paloma Tanner Raul'' and concatenate them.
        
        \textbf{[Answer 1]} 1. Write down the words: Billie Paloma Tanner Raul
        
2. Take the last letter of the first word "Billie", which is "e".

3. Take the last letter of the second word "Paloma", which is "a".

4. Take the last letter of the third word "Tanner", which is "r".

5. Take the last letter of the fourth word "Raul", which is "l".

6. Concatenate these letters together: "ealr".

7. The result word is \textcolor{CBOrange}{"ealr"}.

\textbf{[Answer 2]} Sure, let's break it down step by step:

1. The last letter of "Billie" is "e".

2. The last letter of "Paloma" is "a".

3. The last letter of "Tanner" is "r".

4. The last letter of "Raul" is "l".

Now, if we concatenate these letters, we get the word "earl". Therefore, the answer is \textcolor{CBBlue}{"earl".}

\textbf{[Final answer]} \textcolor{CBBlue}{earl} \\





\midrule

2nd CoT shows a different reasoing pattern and yields the correct answer. & \textbf{Question}: Cornelia likes to travel. She visited already 42 different countries. 20 of them were in Europe and 10 in South America. From the rest of the countries, only half of them were in Asia. How many Asian countries has Cornelia visited?
 
 \textbf{[Answer 1]} 1. We know that Cornelia visited 42 countries in total.
        
2. We also know that 20 of those countries were in Europe and 10 were in South America.

 \textcolor{CBOrange}{3. Therefore, the remaining countries must be half of the total number of countries Cornelia visited, which is 42/2 = 21.}

\textcolor{CBOrange}{4. We know that Cornelia visited only half of the Asian countries, which means there are 21 - 10 = 11 Asian countries left to be determined.}

\textcolor{CBOrange}{5. Therefore, Cornelia visited 11 Asian countries in total.}

\textbf{[Answer 2]} Step 1: Let's start by finding out how many countries Cornelia visited in Europe and South America. We know that she visited 20 countries in Europe and 10 countries in South America, so the total number of countries she visited in these two regions is:
20 + 10 = 30

\textcolor{CBBlue}{Step 2: We know that Cornelia visited 42 countries in total. We also know that 30 of these countries were in Europe and South America. Therefore, the number of countries she visited outside of these regions is: 42 - 30 = 12}

\textcolor{CBBlue}{Step 3: We are told that only half of the countries Cornelia visited outside of Europe and South America were in Asia. This means that the number of countries she visited in Asia is:
12 / 2 = 6}

\textcolor{CBBlue}{Therefore, Cornelia visited 6 Asian countries.}

\textbf{[Final answer]} 6 \\
\bottomrule
    \end{tabular}
    \caption{Examples of self-correction. \textcolor{CBBlue}{Correct reasonings in sky blue.} \textcolor{CBOrange}{Incorrect reasonings in odark range.}}
    \label{table:human_analysis_example}
    \end{center}
\end{table*}

\section{DCoT@3 Answer Patterns}
\label{app:dcot3_ans_patterns}
\Cref{tab:dcot_3_analysis} shows the answer patterns of LLaMA 2 7B with \texttt{DCoT@3} on multiple choice tasks.

\section{DCoT Best $k$ Parameter}
\label{app:bestk}
\Cref{tab:bestk} shows the best $k$ (i.e., number of CoTs) per model and dataset according to the dev set.


\section{Data Generation}
\label{app:data_generation}
We report the CoT triggers used to generate the training CoTs in \Cref{tab:cot_triggers}. As we can see, the triggers condition LLMs to generate answers in different scenarios. This, combined with high-temperature values (0.7), allows GPT 3.5 turbo to generate diverse CoTs. To extract the answers from the CoTs, we used the following format: ``\{cot\} Therefore, the answer (A, B, C, or D) is:'' where we change (A, B, C, D) for the corresponding options of the dataset. If the dataset expects a number and not a list of options, we don't give any list of options in the prompt and extract the number with a regular expression. Lastly, for the span extraction datasets, we use the following template: ``\{text\} \{question\} Answer: \{answer\} \{cot\_trigger\}.'' The idea behind this template is to provide the golden answer and prompt the model to generate rationales that explain that answer and use them as CoTs as in \citep{kim-etal-2023-cot}. The total GPT cost to generate the CoTs is $\$43.68$.

\begin{table*}[t]
\begin{center}
\begin{tabular}{l}
\toprule
\textbf{CoT Triggers} \\\midrule
Answer: Let's think step by step.                            \\
Answer: Before we dive into the answer,                      \\
Answer: Let's think like a detective step by step.           \\
Answer: Let's think about this logically.                    \\
Answer: Let's solve this problem by splitting it into steps. \\
Answer: The answer is after the proof.                       \\
Answer: Let's differentiate using step by step reasoning .   \\
Answer: Let's think step by step using inductive reasoning.  \\
Answer: Let's be concise and think step by step.             \\
Answer: Let's reflect on each answer option step by step.    \\
Answer: Let's think step by step given every option equal consideration.               \\
Answer: Let's think step by step like a scientist.           \\
Answer: Let's use step by step inductive reasoning.          \\
Answer: Let's work by elimination step by step.              \\
Answer: Let's use step by step deductive reasoning.          \\
Answer: Let's work this out in a step by step way to be sure we have the right answer.\\\toprule
because of the following reasons:\\
Justification:\\
Here's why:\\
Here is a list of the reasons:\\
Now, let's think step by step about the reasons:\\
\bottomrule
\end{tabular}
\caption{List of CoT triggers used to generate the training CoTs. The bottom part of the table are the triggers for span extraction datasets.}
\label{tab:cot_triggers}
\end{center}
\end{table*}

\begin{table*}[]
\begin{center}
\begin{tabular}{@{}lcccc@{}}
\toprule
\textbf{Dataset} & \textbf{Phi 1.5} & \textbf{Phi 2} & \textbf{LLaMA2 7B} & \textbf{LLaMA2 13B} \\ \midrule
ARC           & 4 & 2 & 4 & 4 \\
BGQA          & 2 & 4 & 4 & 2 \\
ConditionalQA & 2 & 4 & 4 & 1 \\
GSM8K         & 2 & 3 & 2 & 3 \\
HotpotQA      & 2 & 2 & 1 & 3 \\
LCC           & 1 & 2 & 1 & 2 \\
Quartz        & 4 & 4 & 2 & 2 \\
StrategyQA    & 3 & 1 & 1 & 1 \\ \bottomrule
\end{tabular}
\caption{Best number of cots ($k$ parameter) for each model and dataset in our best DCoT models according to the dev set.}
\label{tab:bestk}
\end{center}
\end{table*}

\end{document}